\documentclass{article}

\usepackage{arxiv}

\usepackage[utf8]{inputenc} 
\usepackage[T1]{fontenc}    
\usepackage{hyperref}       
\usepackage{url}            
\usepackage{booktabs}       
\usepackage{amsfonts}       
\usepackage{nicefrac}       
\usepackage{microtype}      
\usepackage{lipsum}		
\usepackage{graphicx}
\usepackage{natbib}
\usepackage{doi}
\usepackage{amsmath}

\usepackage{epstopdf}

\title{Multi-modal Representation Learning for Cross-modal Prediction of Continuous Weather Patterns from Discrete Low-Dimensional Data}

\author{ \href{https://orcid.org/0009-0008-0706-2913}{\includegraphics[scale=0.06]{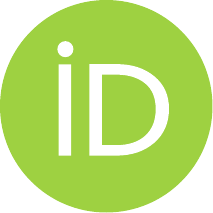}\hspace{1mm}Alif Bin Abdul Qayyum}\\
	Department of Electrical \& Computer Engineering\\
	Texas A\&M University\\
	College Station, TX 77843-3128 \\
	\texttt{alifbinabdulqayyum@tamu.edu} \\
	\And
	\href{https://orcid.org/0000-0002-7621-940X}{\includegraphics[scale=0.06]{orcid.pdf}\hspace{1mm}Xihaier Luo} \\
	Computational Science Initiative\\
    Brookhaven National Laboratory\\
	Upton, NY 11973-5000 \\
	\texttt{xluo@bnl.gov} \\
    \And
    \href{https://orcid.org/0000-0002-2264-3512}{\includegraphics[scale=0.06]{orcid.pdf}\hspace{1mm}Nathan M. Urban} \\
	Computational Science Initiative\\
    Brookhaven National Laboratory\\
	Upton, NY 11973-5000 \\
	\texttt{nurban@bnl.gov} \\
    \And
    \href{https://orcid.org/0000-0002-4347-2476}
    {\includegraphics[scale=0.06]{orcid.pdf}\hspace{1mm}Xiaoning Qian}\thanks{Also affiliated with Computational Science Initiative,
    Brookhaven National Laboratory.} \\
	Department of Electrical \& Computer Engineering\\
	Texas A\&M University\\
	College Station, TX 77843-3128 \\
	\texttt{xqian@tamu.edu} \\
    \And
    \href{https://orcid.org/0000-0001-9328-1101}
    {\includegraphics[scale=0.06]{orcid.pdf}\hspace{1mm}Byung-Jun Yoon}\thanks{Also affiliated with Computational Science Initiative,
    Brookhaven National Laboratory.} \\
	Department of Electrical \& Computer Engineering\\
	Texas A\&M University\\
	College Station, TX 77843-3128 \\
	\texttt{bjyoon@tamu.edu} \\
}

\date{}

\renewcommand{\shorttitle}{\textit{arXiv} Template}

\hypersetup{
pdftitle={Multi-modal Representation Learning for Cross-modal Prediction of Continuous Weather Patterns from Discrete Low-Dimensional Data},
pdfsubject={cs.LG, cs.CV},
pdfauthor={Alif Bin Abdul Qayyum, Xihaier Luo, Nathan M. Urban, Xiaoning Qian, Byung-Jun Yoon},
pdfkeywords={Multi-modal representation learning, continuous super-resolution, dimensionality reduction, cross-modal prediction},
}

\def\x{{\mathbf x}}

\begin{document}
\maketitle

\begin{abstract}
	World is looking for clean and renewable energy sources that do not pollute the environment, in an attempt to reduce greenhouse gas emissions that contribute to global warming. Wind energy has significant potential to not only reduce greenhouse emission, but also meet the ever increasing demand for energy. To enable the effective utilization of wind energy, addressing the following three challenges in wind data analysis is crucial. Firstly, improving data resolution in various climate conditions to ensure an ample supply of information for assessing potential energy resources. Secondly, implementing dimensionality reduction techniques for data collected from sensors/simulations to efficiently manage and store large datasets. Thirdly, extrapolating wind data from one spatial specification to another, particularly in cases where data acquisition may be impractical or costly. We propose a deep learning based approach to achieve multi-modal continuous resolution wind data prediction from discontinuous wind data, along with data dimensionality reduction. 
\end{abstract}



\section{Introduction}
\label{sec:intro}
Human-induced global warming of $1.1^{\circ}$ C has resulted unprecedented change in the earth’s climate in recent human history. Intergovernmental Panel on Climate Change (IPCC) portrays grim future that we may see warming of $3^{\circ}$ C or more by the end of the century, paving ways to disastrous repercussions for ecosystems, economies and human socities, if precautionary measures are not taken immediately to limit the rapidly deteriorating condition of climate \cite{IPCC_2022_WGIII, accelerating-extinct-risk, disease-climate-change, extinct-risk, species-rapid-change}. Humankind needs to shift to renewable energy sources, such as wind power due to its abundance, wide distribution, and no greenhouse gas emission while in operation. \newline
While the advantages of wind energy are evident, attaining its optimal utilization is a formidable endeavor: (1) \textit{Resolution}: Identifying the ideal sites for wind turbines demands a resolution as fine as 1 square kilometer~(km) or even finer. However, most wind simulations struggle to offer such a resolution. (2) \textit{Storage}: Mounting evidence suggests a long-term temporal correlation in weather patterns. Unfortunately, many existing storage systems are limited to accommodating just one month's worth of data, posing challenges for the analysis of wind patterns over multiple years to address seasonal variations. (3) \textit{Acquisition}: Establishing wind measurement stations in specific areas can pose challenges due to the high expenses associated with transportation and maintenance. As a result, to speed up the utilization of wind energy, there is a pressing need to estimate continuous wind pattern from reduced low  dimensional, discontinuous data; and also achieve this in a cross modal fashion where we can estimate wind pattern at inaccessible or expensive spaces from available data at accessible spaces. \newline
Inaccessible and expensive data acquisition procedures at certain space and conditions necessitate multi-modal learning, extrapolation and analysis for wind data. Cross modal inference in wind data can pave the way to cost-effective solutions for data analysis at risky scenarios. A reliable multi-modal model, with accurate prediction capabilities at sufficient high resolution and with data reduction capabilities, can use data at accessible space and conditions acquired through cheap procedures for analysis and decision making on wind data at cases where data is hard to acquire.\newline 
Deep learning based multi-modal learning has made tremendous improvements in multimedia, biology, weather forecasting and many other fields \cite{mm-object-recognition, glue, mmdiffusion, HurricaneForecasting}. Continuous super resolution, a widely studied topic among the computer vision research community \cite{liif, edsr, idm}, has been actively tested on weather data with the help of deep learning\cite{adverserial-sr, wind-solar-sr-benchmark, fourier-downscaling}. Recently, Conditional encoding of low resolution data and Fourier encoding of co-ordinates has achieved greater precision for continuous super resolution in weather patterns \cite{xihaier}. Neural field based continuous super resolution for image data has also been gaining popularity \cite{neural-field, neral-operator-sr}.
Neural network based data reduction models have been studied even before the resurgence of deep learning, commonly termed as autoencoders \cite{autoencoder, anomaly-autoencoder, iunets}. In fact, deep learning based autoencoders have been applied to find hidden patterns in high resolution wind data to assess potential energy resources \cite{find-pattern-wind-flow}. \newline
We propose a deep learning based method for multi-modal continuous wind data pattern reconstruction from reduced low-dimensional discontinuous wind data. A dimension reducing convolutional neural network based encoder reduces the high dimensional wind data into a low dimensional space, not only in a intra-modality fashion, but also in a inter-modality fashion. A coordinate based decoder, inspired by the ideas of neural field \cite{neural-field}, works as a mapping from the low dimensional discontinuous representation to a continuous super resolution space. Together, the overall method maps high resolution discontinuous wind data to a low resolution discontinuous space and then maps to a continuous super-high resolution space, both in intra- \& inter-modality.







\section{Methodology}
\label{sec:methodology}
\begin{figure*}[tbp]
\begin{center}
    \includegraphics[width=\textwidth]{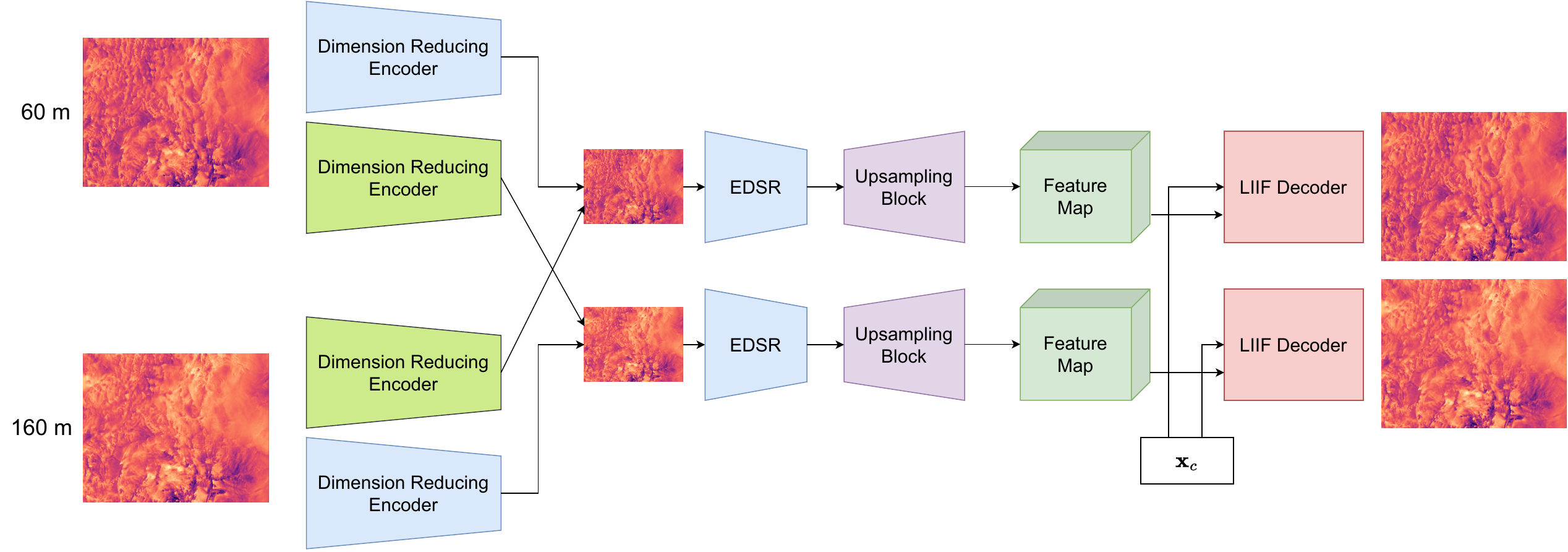}
    \caption{Illustration of the overall model architecture.}
    \label{fig:methodology}
\end{center}
\end{figure*}
The overall methodology has three distinct deep learning models for each modality. They are dimension reducing encoders to encode high resolution weather data into a low resolution space, feature encoders that learn the spatial features from the low resolution representations, and coordinate based decoders that use the extracted features by the feature encoders in the vicinity of a coordinate and predicts the wind data for that specific coordinate. In this work, we worked with 2 different modalities, $\mathcal{M}_0$ and $\mathcal{M}_1$. Figure \ref{fig:methodology} summarizes the proposed methodology.
\subsection{Multi-modal Weather Data}
\label{subsec:mm-weather-data}
We define each modality as wind data at different heights from the ground. Specifically, weather data at $h_0$ units above from the ground is considered as modality $\mathcal{M}_0$, similarly $\mathcal{M}_1$ constitutes data at $h_1$ units above the ground, conditioned on $h_0 \neq h_1$. A detailed description is provided in Section \ref{subsubsec:dataset}.
\subsection{Dimension Reducing Encoder}
\label{subsec:dim-red-enc}
At the beginning of the overall framework, there are several convolutional neural network based dimension reduction encoders that encode the high resolution data to a low resolution space. We have two different kinds of dimension reduction encoders, self-encoders and cross-modal encoders, with the similar architecture. Self encoders convert the high resolution data from one modality to its corresponding low resolution representation, whereas the cross-modal encoders convert the high resolution data from one modality to the low resolution representation of a different modality. Let $\mathcal{M}_0^H \in \mathbb{R}^{c_H \times h_H \times w_H}$ and $\mathcal{M}_1^H \in \mathbb{R}^{h_H \times w_H}$ be the high resolution data space of $\mathcal{M}_0$ and $\mathcal{M}_1$ correspondingly. Similarly $\mathcal{M}_0^L \in \mathbb{R}^{c_L \times h_L \times w_L}$ and $\mathcal{M}_1^L \in \mathbb{R}^{c_L \times h_L \times w_L}$ be the low resolution data space of $\mathcal{M}_0$ and $\mathcal{M}_1$ correspondingly. Let $c_H$, $h_H$ and $w_H$ denote channel depth, height and width of the high resolution data dimension for each modality, $c_L$, $h_L$ and $w_L$ channel depth, height and width of the low resolution data dimension for each modality. $\mathbf{E}_0^{0}:\mathcal{M}_0^H \to \mathcal{M}_0^L$, $\mathbf{E}_1^{1}:\mathcal{M}_1^H \to \mathcal{M}_1^L$
are the self encoders and $\mathbf{E}_0^{1}:\mathcal{M}_0^H \to \mathcal{M}_1^L$, $\mathbf{E}_1^{0}:\mathcal{M}_1^H \to \mathcal{M}_0^L$ are the cross-modal encoders. The architecture of the encoder is inspired from the architecture proposed in the downsampling part of the invertible UNet~\cite{iunets}.
\subsection{Local Implicit Image Function based Decoder}
\label{subsec:liif-dec}
Local implicit image function (LIIF) based decoder is a coordinate based decoding approach which takes the coordinate and the deep features around that coordinate as inputs and outputs the value for that corresponding coordinate \cite{liif}. LIIF-based decoder consists of two distinct models: a residual convolutional neural network based feature encoder, EDSR \cite{edsr}, and a coordinate based decoder. Due to the continuous nature of spatial coordinates, LIIF-based decoder can decode into arbitrary resolution. $\mathbf{FE}_0:\mathcal{M}_0^L \in \mathbb{R}^{c_L \times h_L \times w_L} \to \mathcal{M}_0^F \in \mathbb{R}^{c_F \times h_F \times w_F}$, $\mathbf{FE}_1:\mathcal{M}_1^L \in \mathbb{R}^{c_L \times h_L \times w_L} \to \mathcal{M}_1^F \in \mathbb{R}^{c_F \times h_F \times w_F}$ are two EDSR-based feature encoders for modalities $\mathcal{M}_0$ and $\mathcal{M}_1$ into the encoded feature space $\mathcal{M}_0^F$ and $\mathcal{M}_1^F$ respectively. We use $c_F$, $h_F$, $w_F$ to denote channel depth, height and width of the corresponding encoded feature space. Let $\mathbf{X}_c$ be a 2-D coordinate space. Decoders are functions that take the encoded feature and coordinate as input. $\mathbf{D}_0:\mathbf{x} \in \mathbb{R}^2, \mathcal{M}_0^F \in \mathbb{R}^{c_F \times h_F \times w_F} \to \mathbf{X}^C \in \mathbb{R}$, $\mathbf{D}_1:\mathbf{x} \in \mathbb{R}^2, \mathcal{M}_1^F \in \mathbb{R}^{c_F \times h_F \times w_F} \to \mathbf{X}^C \in \mathbb{R}$ are two coordinate based decoders for modalities $\mathcal{M}_0$ and $\mathcal{M}_1$.
\subsection{Self \& Cross Modality Prediction}
\label{subsec:self-cross-pred}
Let $\mathbf{X}_0^H \in \mathcal{M}_0^H$ be a data instance with high resolution in modality $\mathcal{M}_0$. Our goal is to achieve data reduction and continuous superresolution of this data instance, not only in modality $\mathcal{M}_0$ but also in modality $\mathcal{M}_1$. With the self-encoder $\mathbf{E}_0^0$ and cross-modal encoder $\mathbf{E}_0^1$ we can get the low dimensional representation of this data instance in both modalities, and consequently achieve continuous super resolution with the LIIF based decoder in both modalities, $(\mathbf{FE}_0, \mathbf{D}_0), (\mathbf{FE}_1, \mathbf{D}_1)$. For example, for a co-ordinate point $\mathbf{x}_c \in \mathbf{X}_c$, $\mathbf{D}_0(\mathbf{FE}_0(\mathbf{E}_0^0(\mathbf{X}_0^H)), \mathbf{x}_c)$ represents the prediction at modality $\mathcal{M}_0$ or self-prediction and $\mathbf{D}_1(\mathbf{FE}_1(\mathbf{E}_0^1(\mathbf{X}_0^H)), \mathbf{x}_c)$ represents the prediction at modality $\mathcal{M}_1$ or cross-prediction. Similarly, for a data instance $\mathbf{X}_1^H \in \mathcal{M}_1^H$ and for a co-ordinate point $\mathbf{x}_c \in \mathbf{X}_c$, $\mathbf{D}_1(\mathbf{FE}_1(\mathbf{E}_1^1(\mathbf{X}_1^H)), \x_c)$ represents the prediction at modality $\mathcal{M}_1$ or self-prediction and $\mathbf{D}_0(\mathbf{FE}_0(\mathbf{E}_1^0(\mathbf{X}_1^H)), \x_c)$ represents the prediction at modality $\mathcal{M}_0$ or cross-prediction.
\section{Results}
\label{sec:results}
\begin{figure*}[htb]
\label{fig:results}
\begin{minipage}[b]{\textwidth}
  \centering
  \centerline{\includegraphics[width=\textwidth]{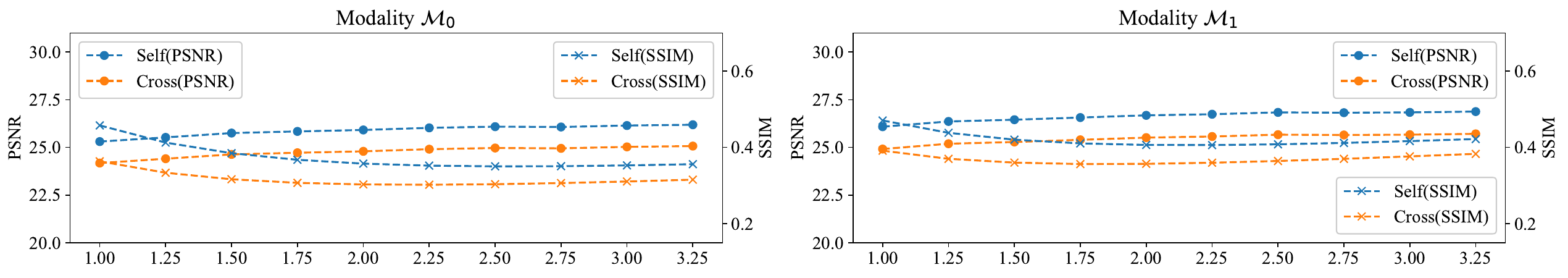}}
  \centering{(a) Average of PSNR and SSIM over test set for northern projection}
\end{minipage}
\begin{minipage}[b]{\textwidth}
  \centering
  \centerline{\includegraphics[width=\textwidth]{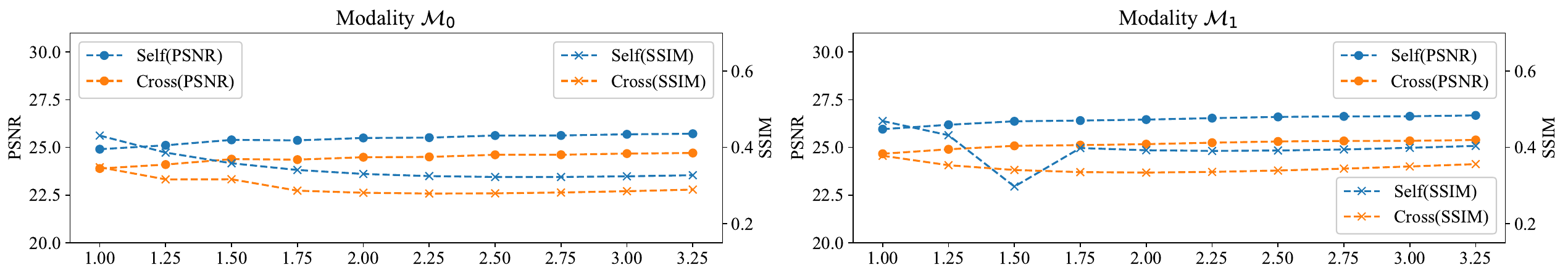}}
  \centering{(b) Average of PSNR and SSIM over test set for eastern projection}
\end{minipage}
\begin{minipage}[b]{0.4975\textwidth}
  \centering
  \centerline{\includegraphics[width=\textwidth]{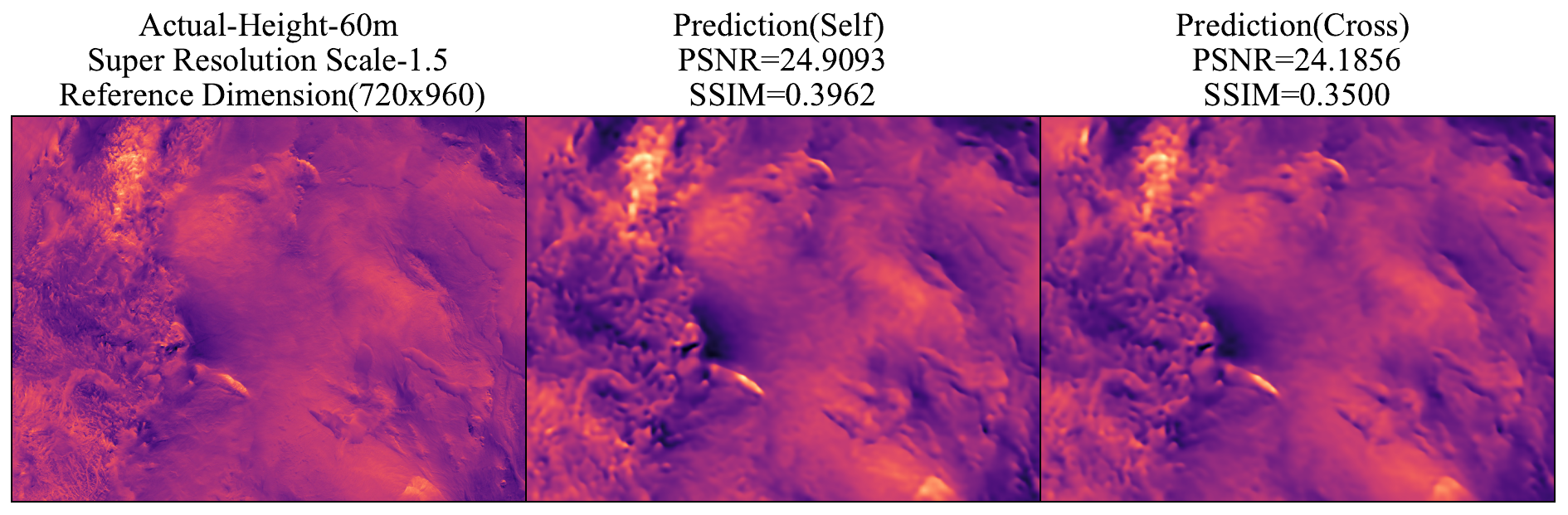}}
\end{minipage}
\begin{minipage}[b]{0.4975\textwidth}
  \centering
  \centerline{\includegraphics[width=\textwidth]{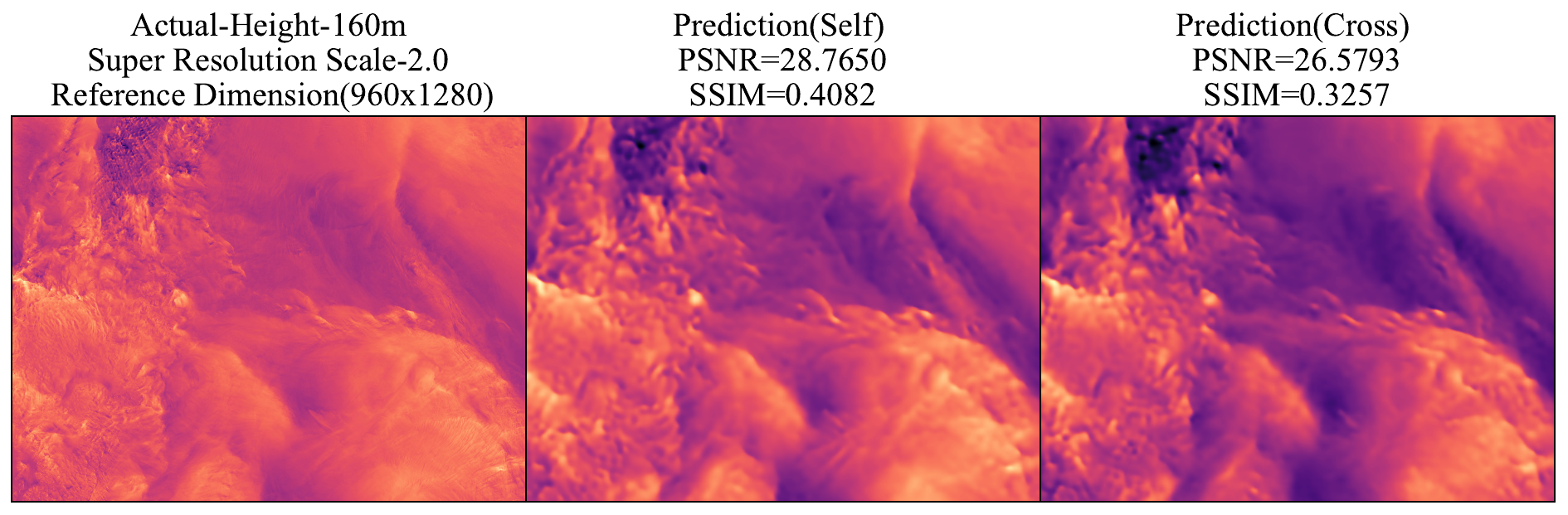}}
\end{minipage}
\begin{minipage}[b]{\textwidth}
  \centering{(c) Some representative results from self and cross predictions at different heights and at different super resolution scale}
\end{minipage}
\caption{Experimental results.}
\end{figure*}
We first introduce the wind data set that we used to qualitatively and quantitatively evaluate our proposed model 
\subsection{Wind Data}
\label{sec:wind-data}
National Renewable Energy Laboratory's Wind Integration National Database (WIND) Toolkit provides high spatial and temporal resolution wind power, wind power forecasting, and meteorological data for over 126,000 locations across the continental United States during a 7-year span \cite{wind-toolkit}. The simulated forecasts were developed using the Weather Research and Forecasting Model, which operates on a 2-kilometer (km) by 2-kilometer (km) grid with a $10$-meter($m$) resolution from the ground to $200m$ above ground with several temporal resolutions available at $1-$hour, $4-$hour, $6-$hour, and day-ahead forecast horizons. Wind velocity data with northern $(u)$ and eastern $(v)$ wind components are used in the following experiments. Wind velocities are determined using the wind speed and direction at $100$ meters. The spatial resolution of the WIND Toolkit is $2 \text{km} \times 1 \text{hr}$ in spatio-temporal resolution. As a result, the wind data set is $1602 (\text{latitude}) \times 2976 (\text{longitude}) \times 61368 (\text{number of instances})$, or almost $1.2$ TB per wind component~(wind data at different heights). We randomly cropped to reduce the resolution to $1500 (\text{latitude}) \times 2000 (\text{longitude})$ for each time instance. 
\subsection{Experimental Setup}
\label{sec:experimental-setup}
\subsubsection{Data Set}
\label{subsubsec:dataset}
We built the data set for multi-modal super-resolution tasks using simulated wind data. We randomly sampled $500$ data points from different timestamps among the total available $61368$ instances for each height above from the ground~($60m$ and $160m$), with $400$ data points for training the models and $100$ data points for testing. We considered wind data at heights $60m$ and $160m$ as the two different modalities. We used bicubic interpolation to generate a pair of high-resolution and super-high-resolution samples for each instance. For example, if the input dimension at both modalities is $(480 \times 640)$, and the super resolution scale is $1.5 \times$, then the output super-high-resolution dimension is $(720 \times 960)$. The process of creating this pair of high and super-high resolution samples is slightly different at train and test phases. For example, while training, at $1.5 \times$ super resolution scale, the super-high resolution sample with dimension $(720 \times 960)$ is created by randomly cropping from the actual $(1500 \times 2000)$ resolution data, and then the high resolution sample with dimension of $(480 \times 640)$ is created by bicubic interpolation from the super-high resolution counterpart. We set the dimension for high resolution data to be $(480 \times 640)$ and varied the super resolution scale from $1 \times$ to $3.25 \times$. So the highest dimension for super-high resolution is $(1500 \times 2000)$, which is same as the actual sampled data from the WIND toolkit. While testing, both the super-high and high resolution samples of a pair is created by bicubic interpolation from the actual $(1500 \times 2000)$ dimension data.
\subsubsection{Training Details}
\label{subsubsec:training-details}
Adam is adopted as the optimizer with a weight decay $\lambda = 0.0001$ \cite{adam}. All the $8$ different sections of the model ($2$ dimension reducing self encoders, $2$ dimension reducing cross encoders, $2$ feature encoders, and $2$ decoders) are trained for $2500$ epochs with an initial learning rate of $10^{-5}$ that decays by $\gamma = 0.9999$ every epoch. At super resolution scale $s$, the super resolution ground truth $\mathbf{X}^S_{\mathcal{M}}$ has a dimension of $(c_H \times s\cdot h_H \times s\cdot w_H)$. The loss function at equation \ref{eq:loss}, a sum of two reconstruction loss terms and a latent loss, is optimized during training. Equation \ref{eq:self-loss} and \ref{eq:cross-loss} defines the reconstruction loss for self and cross modality predictions accordingly. A latent loss function, defined in equation \ref{eq:latent-loss}, is introduced to enforce the predicted low-dimensional representations, both from self and cross encoders to be the same for corresponding modalities. 
\begin{equation}
\label{eq:self-loss}
\begin{split}
    \mathcal{L}_{self} = & MSE(\mathbf{D}_0(\mathbf{FE}_0(\mathbf{E}_0^0(\mathbf{X}_0^H))), \mathbf{X}_0^{S}) \\ & + MSE(\mathbf{D}_1(\mathbf{FE}_1(\mathbf{E}_1^1(\mathbf{X}_1^H))), \mathbf{X}_1^{S})
\end{split}
\end{equation}
\begin{equation}
\label{eq:cross-loss}
\begin{split}
    \mathcal{L}_{cross} = & MSE(\mathbf{D}_0(\mathbf{FE}_0(\mathbf{E}_0^1(\mathbf{X}_1^H))), \mathbf{X}_0^{S}) \\ & + MSE(\mathbf{D}_1(\mathbf{FE}_1(\mathbf{E}_0^1(\mathbf{X}_0^H))), \mathbf{X}_1^{S})
\end{split}
\end{equation}
\begin{equation}
\label{eq:latent-loss}
\begin{split}
    \mathcal{L}_{latent} = & MSE\left(\mathbf{E}_0^0(\mathbf{X}_0^H), \dfrac{\mathbf{E}_0^0(\mathbf{X}_0^H) + \mathbf{E}_1^0(\mathbf{X}_1^H)}{2}\right) \\
    & + MSE\left(\mathbf{E}_1^0(\mathbf{X}_1^H), \dfrac{\mathbf{E}_0^0(\mathbf{X}_0^H) + \mathbf{E}_1^0(\mathbf{X}_1^H)}{2}\right) \\
    & + MSE\left(\mathbf{E}_1^1(\mathbf{X}_1^H), \dfrac{\mathbf{E}_1^1(\mathbf{X}_1^H) + \mathbf{E}_0^1(\mathbf{X}_0^H)}{2}\right) \\
    & + MSE\left(\mathbf{E}_0^1(\mathbf{X}_0^H), \dfrac{\mathbf{E}_1(\mathbf{X}_1^H) + \mathbf{E}_0^1(\mathbf{X}_0^H)}{2}\right)
\end{split}
\end{equation}
\begin{equation}
\label{eq:loss}
\begin{split}
    \mathcal{L} = \mathcal{L}_{self} + \mathcal{L}_{cross} + \mathcal{L}_{latent}
\end{split}
\end{equation}
\subsubsection{Evaluation Metrics}
\label{subsubsec:metrics}
We employed two additional metrics to evaluate the model performance in addition to mean squared error~(MSE) loss between target and predicted wind speed projections. Peak signal-to-noise ratio (PSNR) is the ratio of a signal’s maximum possible value (power) to the power of distorting noise that affects the quality of its representation. Structural similarity index (SSIM) is a perceptual metric that evaluates the degradation of image quality, that compares the spatial structures between the target image and reproduced image. 
\subsubsection{Obtained Results}
We tested the performance of our model at different super resolution scales for both self and cross predictions on the test dataset consisting $100$ datapoints. The high resolution input dimension was set to $(1 \times 480 \times 640)$ and the low resolution representation had a dimension of $(2 \times 60 \times 80)$. Figure \ref{fig:results} shows the average of the evaluation metrics over the test set for both northern and eastern projections. Some predictions at different super resolution scales and at different heights with their corresponding PSNR and SSIM values are also provided. The results show a trend of better performance at self predictions compared to cross predictions. SSIM values tend to go lower as the super resolution scale goes higher. On the contrary, the PSNR values tend to go high with higher super resolution scale, as expected with our model construct.
\section{Conclusion}
\label{sec:conclusion}
In this work, we proposed a novel coordinate-based deep learning solution for achieving continuous super-resolution, data dimensionality reduction, and multi-modal learning of climatological data, all three at the same time. We specifically developed a local implicit neural network model for learning continuous, rather than discrete, representations of climate data, such as wind velocity fields used for wind farm power modeling across the continental United States, along with multi-modal dimension reducing encoder that facilitates dimension reduction and cross modality extrapolation. We also introduced a latent loss function to the optimization procedure to ensure cross modality learning. Obtained results have shown the promising potential to solve real-life scenarios in wind energy resource assessment for electricity generation, efficient storage of huge amount of data by dimensionality reduction and extrapolation of data to inaccessible spatial spaces from available wind data.
\section*{Acknowledgement}
This work was supported in part by the Department of Energy (DOE) award DE-SC0012704.

\bibliographystyle{unsrtnat}
\bibliography{template}  






\end{document}